\documentclass[letterpaper]{article} 
\usepackage{aaai25}  
\usepackage{times}  
\usepackage{helvet}  
\usepackage{courier}  
\usepackage[hyphens]{url}  
\usepackage{graphicx} 
\usepackage{natbib}  
\usepackage{caption} 
\usepackage{algorithm}
\usepackage{algorithmic}

\usepackage{multirow}
\usepackage{graphicx}
\usepackage{epstopdf}
\usepackage{booktabs}
\usepackage{amsmath}
\usepackage{verbatim}
\usepackage{placeins}
\usepackage{xcolor} 
\usepackage{makecell}
\usepackage{amssymb}
\definecolor{cvprblue}{rgb}{0.21,0.49,0.74}
\usepackage{newfloat}
\usepackage{listings}
\DeclareCaptionStyle{ruled}{labelfont=normalfont,labelsep=colon,strut=off} 
\floatstyle{ruled}
\newfloat{listing}{tb}{lst}{}
\floatname{listing}{Listing}
\title{Beyond Human Data: Aligning Multimodal Large Language Models \\ by Iterative Self-Evolution}
\author{
    Wentao Tan\textsuperscript{\rm 1,2}~~~~~
    Qiong Cao\textsuperscript{\rm 2}\thanks{Project Lead.}\footnotemark[2]~~~~~
    Yibing Zhan\textsuperscript{\rm 2}\thanks{Corresponding Authors.}~~~~~
    Chao Xue\textsuperscript{\rm 2}~~~~~
    Changxing Ding\textsuperscript{\rm 1,3}~~~~~
}
\affiliations{
    \textsuperscript{\rm 1}South China University of Technology \quad
    \textsuperscript{\rm 2}JD Explore Academy, Beijing \quad
    \textsuperscript{\rm 3}Pazhou Lab, Guangzhou
    ftwentaotan@mail.scut.edu.cn ~~ \{caoqiong1, xuechao19\}@jd.com ~~ zybjy@mail.ustc.edu.cn ~~ chxding@scut.edu.cn
}

\usepackage{bibentry}

\begin{document}

\maketitle

\begin{abstract}
Human preference alignment can significantly enhance the capabilities of Multimodal Large Language Models (MLLMs). However, collecting high-quality preference data remains costly. One promising solution is the self-evolution strategy, where models are iteratively trained on data they generate. Current multimodal self-evolution techniques, nevertheless, still need human- or GPT-annotated data. Some methods even require extra models or ground truth answers to construct preference data. To overcome these limitations, we propose a novel multimodal self-evolution framework that empowers the model to autonomously generate high-quality questions and answers using only unannotated images. First, in the question generation phase, we implement an image-driven self-questioning mechanism. This approach allows the model to create questions and evaluate their relevance and answerability based on the image content. If a question is deemed irrelevant or unanswerable, the model regenerates it to ensure alignment with the image. This process establishes a solid foundation for subsequent answer generation and optimization. Second, while generating answers, we design an answer self-enhancement technique to boost the discriminative power of answers. We begin by captioning the images and then use the descriptions to enhance the generated answers. Additionally, we utilize corrupted images to generate rejected answers, thereby forming distinct preference pairs for effective optimization. Finally, in the optimization step, we incorporate an image content alignment loss function alongside the Direct Preference Optimization (DPO) loss to mitigate hallucinations. This function maximizes the likelihood of the above generated descriptions in order to constrain the model's attention to the image content. As a result, model can generate more accurate and reliable outputs. Experiments demonstrate that our framework is competitively compared with previous methods that utilize external information, paving the way for more efficient and scalable MLLMs. The code is available at \textcolor{cvprblue}{\url{https://github.com/WentaoTan/SENA}}.
\end{abstract}

\section{Introduction}

\begin{figure*}[]
\centerline{\includegraphics[width=1.0\linewidth]{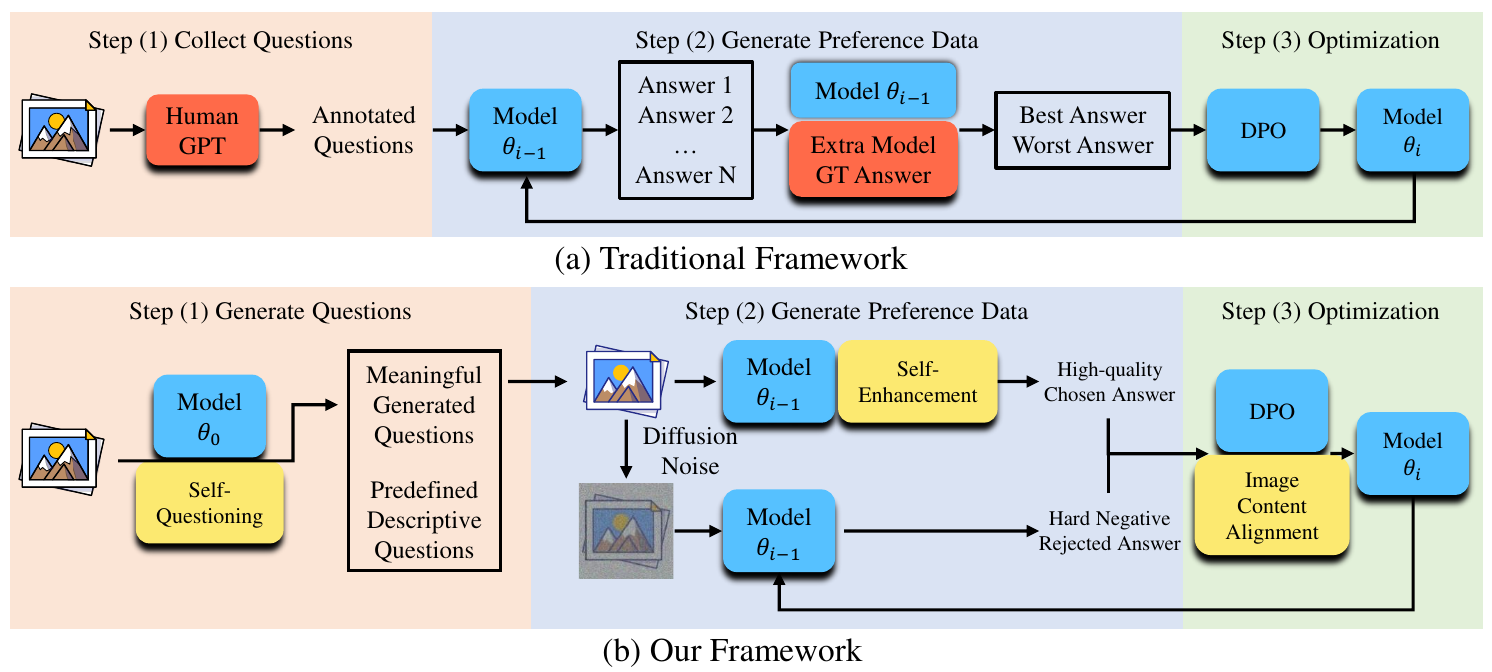}}\caption{
Comparisons between (a) traditional framework and (b) our framework. Our framework combines carefully designed prompt mechanisms and an alignment function, completely eliminating the reliance on annotated data and additional models. 
}
  \label{fig:figure1}
\end{figure*}

An effective strategy for enhancing MLLM capabilities is human preference alignment \cite{amirloo2024understanding,li2023silkie,zhou2024aligning}, which focuses on training models to better align with user preferences using high-quality preference data. This approach can improve the model’s resistance to hallucinations and its ability to follow complex instructions. However, collecting high-quality preference data is labor-intensive and costly, often requiring extensive manual annotation \cite{sun2023aligning,yu2024rlhf} or relying on data generated by advanced models like GPT-4v \cite{achiam2023gpt,li2023silkie,zhou2024aligning}.

To achieve low-cost human preference alignment, self-evolution methods have emerged, enabling models to construct preference data to train themselves \cite{deng2024enhancing,wang2024enhancing,ahn2024srt,yu2024rlaif,zhu2024self}. Fig. \ref{fig:figure1} (a) shows the exact steps of existing self-evolution methods. First, questions annotated by humans or GPT are collected. Then, the evolved model answers. Some methods generate multiple answers and evaluate their quality through a self-rewarding mechanism \cite{yuan2024self} or by using extra MLLMs \cite{yu2024rlaif}, CLIP \cite{rafailov2024direct,zhou2024calibrated}, or ground-truth answers \cite{wang2024enhancing}. Other methods initially generate chosen answers, then create rejected responses by modifying image content \cite{zhu2024self} or using misleading prompts \cite{deng2024enhancing}. 
Finally, a DPO loss function is applied to align with human preference. Despite their promise, current methods often rely on annotated data and additional models, which increases complexity.

To address the above limitations, we explore the potential of going beyond human data and establish a straightforward and efficient multimodal self-evolution framework: \textbf{Only a set of unlabeled images is needed for any model to further improve its performance!}

Our framework, depicted in Fig. \ref{fig:figure1} (b), operates with a single model and unlabeled images, eliminating the need for annotated data by having the model generate both questions and answers. We tackle three significant challenges in this process. The first challenge is generating reliable questions, as meaningless questions lead to useless training data. We introduce an image-driven self-questioning mechanism where the model verifies the answerability of generated questions based on the image content and regenerates them if necessary. This enhances question quality and facilitates effective learning. Additionally, to fully utilize the diverse visual information in the images, we add descriptive questions \cite{liu2024visual} that encourage the model to output descriptions of the images. These descriptions will be utilized to address the other two challenges.

The second challenge is generating discriminative answer pairs. Our experiments show that randomly generated answers often exhibit similar quality, rendering them unsuitable for preference alignment. While some methods \cite{zhu2024self,deng2024enhancing} use corrupted images to create hard negative answers, simply producing less informative rejected answers is insufficient. We propose enhancing discriminative ability by improving the quality of chosen answers through an answer self-enhancement mechanism. The model generates an initial answer and then refines it using the image description, resulting in a more precise chosen answer. For generating rejected answers, we utilize images augmented with diffusion noise. This approach not only enhances data discriminability and ensures robust preference optimization but also allows the model to verify and correct the chosen answers, ensuring response accuracy.

The third challenge is improving the model's resistance to hallucinations. Since the model uses its generated data for self-training, reducing hallucinations is vital. Research shows that models may generate incorrect answers without referencing the actual image content \cite{huang2024opera,jiang2024hallucination,liu2023mitigating,zhou2023analyzing,zhang2020causal,zhang2020feature}. Inspired by this, we propose an image content alignment loss function, which maximizes the likelihood of the generated descriptions to shift the model's attention toward actual image content. By combining this with the DPO loss, our approach ensures that as the model learns user preferences, it remains focused on the actual content of the images, resulting in more accurate outputs.

To the best of our knowledge, this is the first multimodal iterative self-evolution framework that requires no labeled data. We name this framework \textbf{SENA} because it effectively integrates image-driven \textbf{S}elf-questioning, answer self-\textbf{EN}hancement, and image content \textbf{A}lignment. In our framework, the model can generate reliable and discriminative preference data, ensuring stable human preference alignment and continuous performance improvement. Experimental results confirm that our approach significantly enhances the model's performance across multiple benchmarks, encompassing both generative and discriminative tasks.

\begin{figure*}[ht]
\centerline{\includegraphics[width=1.0\linewidth]{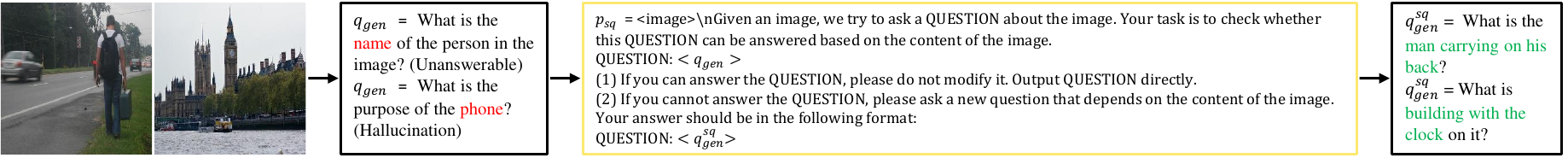}}
  \caption{
 Illustration of the Image-Driven Self-Questioning. SQ checks whether $q_{gen}$ can be answered based on the content of the image. If it cannot, a new question relevant to the image content is generated. The majority of poor-quality questions can be transformed into reliable ones through just one check. Best viewed by zooming in.
}
  \label{fig:figure_sq}
\end{figure*}

\section{Related Work}
\textbf{Self-Evolution in LLMs.}
Self-evolution is initially proposed in the realm of LLMs \cite{calandriello2024human,rosset2024direct,guo2024direct,swamy2024minimaximalist,xiong2024iterative,lu2023self}, also known as iterative DPO or iterative RLHF \cite{dong2024rlhf}. This approach allows models to align with human preferences using their own-generated data, which significantly reduces annotation costs while improving performance \cite{yuan2024self}. A key challenge in self-evolution is constructing reasonable preference data. One method, self-play \cite{chen2024self}, uses open-source supervised fine-tuning (SFT) data, taking ground truth answers as chosen responses and the model's outputs as rejected ones. This process effectively transforms a weak LLM into a strong one. Another method, self-reward \cite{yuan2024self}, employs a small amount of ranking data to train the model to score its own responses, enabling it to generate reasonable preference data with only 5K labeled examples. 

\textbf{Self-Evolution in MLLMs.} This technique is now applied in the MLLM domain \cite{zhou2024calibrated,ahn2024srt,tan2024harnessing,tan2023style,wang2024decoupled,wang2022uncertainty}. For example, RLAIF-V \cite{yu2024rlaif} assesses responses generated by the 7B LLaVA-1.5 model \cite{liu2024improved} using the 34B LLaVA-NEXT model \cite{liu2024llavanext} to obtain preference data pairs. SIMA \cite{wang2024enhancing} classifies responses based on ground truth answers. However, these methods often rely on external models or ground-truth answers, complicating the framework. Other approaches generate chosen answers conventionally and create rejected answers using corrupted images or misleading prompts. For instance, SeVa \cite{zhu2024self} employs images contaminated with diffusion noise to prompt the model to generate responses that differ from the actual content. Moreover, STIC \cite{deng2024enhancing} uses misleading prompts to induce hallucinated responses as rejected answers. Current multimodal methods depend heavily on open-source data, which limits their flexibility and scalability. In contrast, our proposed SENA allows the model to autonomously generate open-ended questions, effectively addressing these limitations and filling a significant gap in the MLLMs.

\section{Method}
Our framework encourages the model to autonomously generate reliable questions and discriminative answers for unlabeled images, iteratively enhancing its capabilities through human preferences alignment. The overall process is summarized in Algorithm \ref{alg:method} and illustrated in Fig. \ref{fig:figure1} (b). Given an initial model $\theta_0$ and a database of images $D$, we plan to evolve $N$ times with $M$ images in each iteration. Thus, we randomly select $N \times M$ images from $D$. 

\subsection{(1) Generate Questions}
Traditional methods rely on human- or GPT-annotated questions, allowing the model only to generate answers. Ideally, the model should be capable of generating both questions and answers simultaneously. Therefore, we use the model $\theta_0$ to generate questions for each sampled image $x$ with the prompt $p_{base}$: \emph{``Please look at the image and generate a question related to the content of the image."}. The generated questions are denoted as $q_{gen} \sim \theta_0 (x, p_{base})$.

\textbf{Image-Driven Self-Questioning.}
While $p_{base}$ prompts the model to generate questions based on the image content, it sometimes produces nonsensical or irrelevant $q_{gen}$, as shown in Fig. \ref{fig:figure_sq}. These flawed questions negatively impact subsequent answers and hinder model optimization. To address this, we introduce an Image-Driven Self-Questioning (SQ) mechanism. The model evaluates whether $q_{gen}$ can be answered based on the image content using the prompt $p_{sq}$. If it cannot, a new question, $q_{gen}^{sq} \sim \theta_{0} (x,q_{gen},p_{sq})$, is generated. This process ensures reliable questions and lays a solid foundation for self-evolution.

Moreover, we find that some $q_{gen}^{sq}$ tend to focus on the prominent objects in the image. To help the model learn about other visual details, we introduce a descriptive question $q_{des}$, which is randomly sampled from a set of generic prompts $P_{des}$ \cite{liu2024visual}. The set $P_{des}$ can be found in the Supplementary Materials, with one example being \emph{``Describe the image concisely."}. 
This addition results in $N \times M$ image-question triplets, represented as $(x, q_{des}, q_{gen}^{sq})$. Once the images and questions are prepared, the model iteratively evolves as described below.

\begin{figure*}[ht]
\centerline{\includegraphics[width=1.0\linewidth]{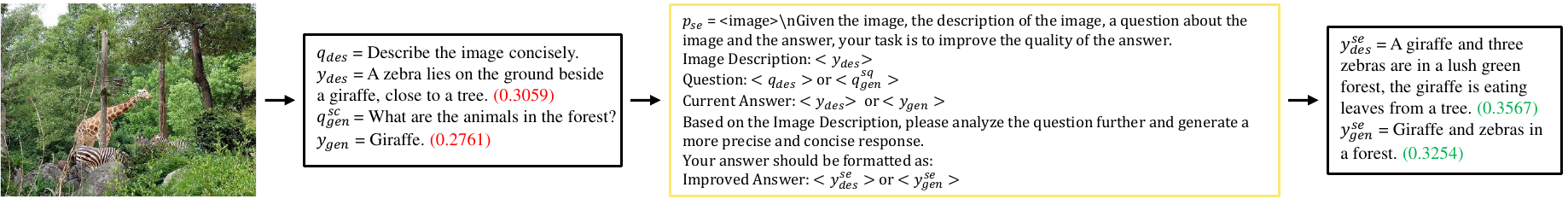}}
  \caption{
 Illustration of the Answer Self-Enhancement techniques. SE analyzes the previous question-and-answer pairs with the help of the image description and enhances the responses. The values in parentheses represent the CLIP scores of the answer-image pairs, which we use to indicate the quality of the answers. Best viewed by zooming in.
}
  \label{fig:figure_se}
\end{figure*}

\subsection{(2) Generate Preference Data} 
At the start of the $i$-th iteration ($1 \leq i \leq N$), the model $\theta_{i-1}$ generates chosen and rejected answers for the image-question dataset. Some methods utilize $\theta_{i-1}$ to randomly generate multiple answers, labeling them as chosen or rejected based on evaluations from self-rewarding mechanisms \cite{yuan2024self}, additional MLLMs \cite{yu2024rlaif}, or ground-truth answers \cite{wang2024enhancing}. However, as shown in the Supplementary Materials, these randomly generated answers often have similar quality, making it challenging to create discriminative preference pairs.

A more effective approach involves using original images for generating chosen answers and content-distorted images for rejected answers \cite{zhu2024self,leng2024mitigating}. For a data point $(x, q_{des}, q_{gen}^{sq})$, the original image $x$ and the question are input into the model to produce the chosen answer $y_w$. The rejected answer $y_l$ is generated using the noisy image $x'$, obtained by adding $T$ times of diffusion noise to $x$. Formally, $y_w \sim \theta_{i-1}(x, q)$ and $y_l \sim \theta_{i-1}(x', q)$, where $q \in \{q_{des}, q_{gen}^{sq}\}$. While this method constructs preference data effectively, it may not yield sufficiently discriminative pairs, hindering the learning efficiency. Although more aggressive image corruption techniques could be employed, they may lead to unstable training \cite{zhu2024self}. 

\textbf{Answer Self-Enhancement.} To address this issue, we focus on enhancing the quality of $y_w$. Fortunately, the answer to question \( q_{des} \) serves as a valuable image description, providing prior knowledge that helps the model better understand questions, analyze existing answers, and ultimately improve answer quality. Therefore, we propose an Answer Self-Enhancement (SE) technique, as detailed in Fig. \ref{fig:figure_se}. For clarity, we denote the answers $y_w$ to \( q_{des} \) and \( q_{gen}^{sq} \) as \( y_{des} \) and \( y_{gen} \), respectively. The SE technique uses \( y_{des} \) to enhance both answers: \( y_{des}^{se} \sim \theta_{i-1}(x, y_{des}, q_{des}, y_{des}, p_{se}) \) and \( y_{gen}^{se} \sim \theta_{i-1}(x, y_{gen}, q_{gen}^{sq}, y_{des}, p_{se}) \). The enhanced answers \( y_{des}^{se} \) and \( y_{gen}^{se} \) become the new chosen answers \( y_{w}^{se} \), along with $y_l$ to create more discriminative preference data. Although it may seem unusual for SE to enhance \( y_{des} \) using itself, this strategy allows the model to reassess the question and generate better answers.
Since each image has two questions, we ultimately generate a total of $2M$ samples for subsequent human preference alignment.

\begin{algorithm}[t]
\caption{SENA: Multimodal Self-Evolution Framework}\label{alg:method}
\begin{algorithmic}[1]

\REQUIRE Image database $D$, initial model $\theta_0$, number of iterations $N$, number of images per iteration $M$, descriptive questions set $P_{des}$

\STATE Randomly sample $N \times M$ images $\{x^k\}_{k=1}^{N \times M}$ from $D$
\STATE \emph{$\#$ (1) Generate Questions}
\FOR{each image $x \in \{x^k\}$}
    \STATE Generate question $q_{gen} \sim \theta_0(x, p_{base})$
    \STATE \textbf{SQ: $q_{gen}^{sq} \sim \theta_{0} (x,q_{gen},p_{sq})$}
    \STATE Randomly sample descriptive question $q_{des} \sim P_{des}$
    \STATE Form triplet $(x, q_{des}, q_{gen}^{sq})$
\ENDFOR

\FOR{$i = 1$ to $N$}
\STATE Use the $i$-th part of $M$ triplets
\STATE \emph{$\#$ (2) Generate Preference Data}
    \FOR{each triplet $(x, q_{des}, q_{gen}^{sq})$ in the $i$-th part}
        \STATE Generate chosen response $y_w \sim \theta_{i-1}(x, q), q \in \{q_{des}, q_{gen}^{sq}\}$
        \STATE Apply diffusion noise to $x$, denote as $x'$
        \STATE Generate rejected response $y_l \sim \theta_{i-1}(x', q), q \in \{q_{des}, q_{gen}^{sq}\}$
        \STATE \textbf{SE: $y_w^{se} \sim \theta_{i-1} (x,y_w,q, y_{des},p_{se})$}
    \ENDFOR
    \STATE \emph{$\#$ (3) Optimization}
    \FOR{each data $(x, y_w^{se}, y_l, q), q \in \{q_{des}, q_{gen}^{sq}\}$}
        \STATE Compute $\pi_{\theta_{i-1}}(y|x, q)$ for $y_w^{se}$ and $y_l$
        \STATE Compute $L_{DPO}$ + $L_{Align}$ \textbf{(CA)} and update
    \ENDFOR
\ENDFOR
\ENSURE $\theta_N$
\end{algorithmic}
\end{algorithm}

\begin{table*}[]
\centering
\setlength{\tabcolsep}{1.0mm} 
\small
\begin{tabular}{c||ccc|c||c|c|cc|cccc|cc|c}
\toprule[1pt]
\multirow{3}{*}{Method} & \multicolumn{3}{c|}{Component} & \multirow{3}{*}{Iteration} & \multicolumn{8}{c|}{Generative Task} & \multicolumn{3}{c}{Discriminative Task} \\ \cline{2-4} \cline{6-16} 
 & \multirow{2}{*}{SQ} & \multirow{2}{*}{SE} & \multirow{2}{*}{CA} &  & \multicolumn{1}{c|}{\multirow{2}{*}{LLaVA$^{\mathrm{W}}$}} & \multicolumn{1}{c|}{\multirow{2}{*}{MM-VET}} & \multicolumn{2}{c|}{MMHal} & \multicolumn{4}{c|}{AMBER-Gen.} & \multicolumn{2}{c|}{AMBER-Dis.} & \multirow{2}{*}{MMBench} \\ 
 &  &  &  &  & \multicolumn{1}{c|}{} & \multicolumn{1}{c|}{} & Score & \multicolumn{1}{c|}{Rate$\downarrow$} & CHAIR$\downarrow$ & Cover & Hal$\downarrow$ & Cog$\downarrow$ & Accuracy & \multicolumn{1}{c|}{F1} &  \\ \hline
$\theta_0$ &&&&&59.6 & 31.7 & 1.90 & 0.61 & 7.6 & \textbf{51.8} & 35.1 & 4.3 & 71.7  & 74.3 & 64.6 \\ \hline
$\theta_1^{Base}$&&&&1& 62.8	&33.7	&1.88	&0.62	&7.4	&51.0	&34.4	&3.4	&70.4	&72.4	&65.1 \\ 
$\theta_1^{SQ}$ &$\checkmark$&&&1&64.9	&33.9	&2.01	&0.59	&6.3	&50.8	&30.4	&3.0	&71.6	&73.9	&\textbf{65.3} \\ 
$\theta_1^{SE}$ &&$\checkmark$&&1&65.0	&34.4	&2.08	&0.56	&6.3	&50.7	&31.4	&2.9	&72.2	&75.1	&\underline{65.2} \\ 
$\theta_1^{CA}$ &&&$\checkmark$&1&66.7&33.2	&2.17	&0.56	&6.5	&51.5	&30.8	&3.5	&74.8	&77.9	&\textbf{65.3} \\ 
\hline
$\theta_1$ &$\checkmark$&$\checkmark$&$\checkmark$&1&66.9	&34.2	&2.28	&0.54	&5.6	&\underline{51.6}	&25.2	&1.9	&75.1	&79.8	&65.0 \\
$\theta_2$ &$\checkmark$&$\checkmark$&$\checkmark$&2&\textbf{67.5}	&\underline{35.1}	&\textbf{2.40}	&\textbf{0.51}	&\underline{5.3}	&50.6	&\underline{23.3}	&\textbf{1.6}	&\underline{75.7}		&\underline{80.2}	&\underline{65.2} \\
$\theta_3$ &$\checkmark$&$\checkmark$&$\checkmark$&3&\underline{67.4}	&\textbf{35.8}	&\underline{2.33}	&\underline{0.52}	&\textbf{4.9}	&49.4	&\textbf{20.5}	&\underline{1.7}	&\textbf{79.4}	&\textbf{83.6}	&64.8 \\
\hline
\end{tabular}
\caption{Ablation study on each key component SQ, SE and CA. $\theta_0$ is the LLaVA-1.5-7B model.}
\label{tab:ablation_study}
\end{table*}

\subsection{(3) Optimization}
We utilize the widely adopted Direct Preference Optimization (DPO) loss to align the model $\theta_{i-1}$ with human preferences. DPO will first construct a reference model $\theta_{ref}$, which is initialized with the parameters from $\theta_{i-1}$ and kept frozen. The goal of DPO is to ensure that as evolution progresses, the current model $\theta_{i-1}$ is more likely to generate high-quality answers $y_w$ than $\theta_{ref}$, and less likely to generate rejected answers $y_l$ compared to $\theta_{ref}$. Specifically, given the input image $x$, question $q$, and output sequence $y$, the likelihood $\pi_{\theta_{ref}}(y|x,q)$ is computed as:
\begin{align}
\pi_{\theta_{ref}}(y|x,q) &= \prod_{s=1}^{|y|} P_{\theta_{ref}}(y | x, q, y_{<s}),
\end{align}
where $|y|$ represents the token length of $y$. The DPO loss function is then defined as:
\begin{align}
L_{DPO} = -\log \sigma \bigg( &\beta \log \frac{\pi_{\theta_{i-1}}(y_w^{se}|x,q)}{\pi_{\theta_{ref}}(y_w^{se}|x,q)} \notag \\
&- \beta \log \frac{\pi_{\theta_{i-1}}(y_l|x,q)}{\pi_{\theta_{ref}}(y_l|x,q)} \bigg),
\end{align}
where $\sigma$ is the sigmoid function, $\beta$ is a hyperparameter that adjusts the loss sensitivity to preference differences, and $q \in \{q_{des}, q_{gen}^{sq}\}$. Notably, although we use the noisy image $x'$ for generating rejected responses, the likelihood is still calculated based on the original image $x$. 

\textbf{Image Content Alignment.} As the model relies on generated data for training, improving its resistance to hallucinations is essential for ongoing evolution. Research indicates that models may produce incorrect answers without referencing actual image content \cite{huang2024opera}. To address this, we design an Image Content Alignment (CA) function to steer the model's focus towards the image content:
\begin{align}
L_{align} = - \frac{1}{|y_{des}^{se}|} \log \pi_{\theta_{i-1}}(y_{des}^{se}|x,q_{des}).
\end{align}
By maximizing the generation probability of the image content \( y_{des}^{se} \), we guide the model's attention to the images, facilitating better understanding and interpretation. As the model's descriptions become more precise, it can produce higher-quality answers through SE, creating a positive feedback loop that continuously improves performance.
Now the final optimization function is given by:
\begin{align} \label{eq:final}
& L_{total} = L_{DPO} + L_{align}.
\end{align} 

Ultimately, we have completed one iteration and advanced the model from $\theta_{i-1}$ to $\theta_i$. SENA will repeat Steps (2) and (3) until $i = N$. The three key designs are complementary; SQ and SE generate high-quality training data that aids CA's learning, while CA enhances the model's generative abilities, enabling SE to function more effectively.

\section{Experiments}
\subsection{Implementation Details}
The dataset $D$ is sourced from the LLaVA665k SFT dataset \cite{liu2024improved}, which includes COCO \cite{lin2014microsoft}, GQA \cite{hudson2019gqa}, TextVQA \cite{singh2019towards}, OCRVQA \cite{mishra2019ocr}, and Visual Genome \cite{krishna2017visual}, totaling 665K images. We conduct three iterations of evolution, setting $N$ = 3. And we plan to use approximately 1\% of $D$ per iteration, amounting to $M$ = 6K images. Only images are used without annotation, resulting in a final random sample of 18K images.

We employ LLaVA-1.5-vicuna-7B \cite{liu2024improved} as the initial model $\theta_0$. All outputs are generated using greedy decoding. Each iteration consists of 1 epoch with a batch size of 128 and a learning rate of 2e-6. The number of diffusion noise additions, \(T\), is set to 600, and the scaling parameter \(\beta\) in DPO is fixed at 0.1.

\begin{table}[]
\centering
\setlength{\tabcolsep}{1.0mm} 
\begin{tabular}{cc||c|ccc}
 \toprule[1pt]
\multicolumn{2}{c||}{Apply SE on} & \multirow{2}{*}{LLaVA$^{\mathrm{W}}$} & \multicolumn{3}{c}{AMBER} \\ \cline{1-2}  
$y_{des}$ & $y_{gen}$ &  & CHAIR$\downarrow$ & Accuracy & F1 \\ \hline
 &  & 62.8 & 7.4 & 70.4 & 72.4 \\ \hline
 $\checkmark$&  & 63.9 & 6.6 & 71.3 & 74.1 \\
 &$\checkmark$  & 63.7 & 6.8 & 72.0 & 74.4 \\
 $\checkmark$&$\checkmark$  & \textbf{65.0} & \textbf{6.3} & \textbf{72.2} & \textbf{75.1} \\ \hline
\end{tabular}
\caption{Impact of applying Self-Enhancement on different answers.}
\label{tab:si_term}
\end{table}

\begin{table}[]
\centering
\setlength{\tabcolsep}{1.0mm}
\begin{tabular}{cc||c|ccc}
 \toprule[1pt]
\multicolumn{2}{c||}{Apply AL on} & \multirow{2}{*}{LLaVA$^{\mathrm{W}}$} & \multicolumn{3}{c}{AMBER} \\ \cline{1-2}  
$y_{des}$ & $y_{gen}$ &  & CHAIR$\downarrow$ & Accuracy & F1 \\ \hline
&  &  62.8 & 7.4 & 70.4 & 72.4 \\ \hline
&$\checkmark$ & 63.3 & 6.5 & 73.5 & 77.1 \\
$\checkmark$&$\checkmark$ & 64.7 & \textbf{6.4} & \textbf{75.0} & \textbf{78.2} \\
$\checkmark$& & \textbf{66.7} & 6.5 & 74.8 & 77.9 \\ \hline
\end{tabular}
\caption{Impact of applying alignment loss (AL) on different answers.}
\label{tab:sa_term}
\end{table}

\begin{table*}[]
\centering
\small
\setlength{\tabcolsep}{1.0mm}
\begin{tabular}{l||c|c|cc|cccc|cc|c}
\toprule[1pt]
\multirow{3}{*}{Method} & \multicolumn{8}{c|}{Generative Task} & \multicolumn{3}{c}{Discriminative Task} \\ \cline{2-12} 
 & \multicolumn{1}{c|}{\multirow{2}{*}{LLaVA$^{\mathrm{W}}$}} & \multicolumn{1}{c|}{\multirow{2}{*}{MM-VET}} & \multicolumn{2}{c|}{MMHal} & \multicolumn{4}{c|}{AMBER-Gen.} & \multicolumn{2}{c|}{AMBER-Dis.} & \multirow{2}{*}{MMBench} \\
 & \multicolumn{1}{c|}{} & \multicolumn{1}{c|}{} & Scores & \multicolumn{1}{c|}{Rate$\downarrow$} & CHAIR$\downarrow$ & Cover & Hal$\downarrow$ & Cog$\downarrow$ & Accuracy  & \multicolumn{1}{c|}{F1} &  \\ \hline \hline
BLIP-2 \cite{li2023blip} &38.1	&22.4	&-	&-	&-	&-	&-	&-	&-	&-		&-\\ 
MiniGPT-4 \cite{zhu2023minigpt}&-	&22.1	&-	&-	&13.6	&63.0	&65.3	&11.3	&63.6	&64.7	&30.9 \\
InstructBLIP-7B \cite{instructblip}&60.9	&26.2	&2.10	&0.58	&8.8	&52.2	&38.2	&4.4	&76.5	&81.7	&38.4\\ 
Shikra-13B \cite{chen2023shikra}  &-	&-	&-		&-	&-	&-	&-	&-	&-	&-	&58.8\\ 
Qwen-VL-7B \cite{bai2023qwen} &60.9	&26.2	&-	&-	&8.8	&52.2	&38.2	&4.4	&76.5	&81.7	&38.4 \\ 
mPLUG-Owl2 \cite{ye2024mplug}  &59.9	&36.2	&-	&-	&10.6	&52.0	&39.9	&4.5	&75.6	&78.5	&63.5 \\

\hline \hline

LLaVA-1.5-7B$^{\dag}$ \cite{liu2024improved} & 59.6 & 31.7 & 1.90 & 0.61 & 7.6 & 51.8 & 35.1 & 4.3 & 71.7 & 74.3 & 64.6 \\ 
\midrule
\multicolumn{6}{l}{\textit{with annotated data or extra models:}} \\
\midrule 
+ SeVa$^{\dag}$ \cite{zhu2024self} &63.3	&\textbf{37.0}	&2.12	&0.57	&7.3	&\textbf{54.0}	&37.3	&2.9	&\underline{79.3}	&\textbf{83.6}	&\textbf{65.6} \\
+ STIC$^{\dag\dag}$ \cite{deng2024enhancing} &63.0	&31.8	&2.07	&0.56	&7.6	&\underline{52.1}	&35.8	&4.4	&71.6	&74.2	&64.3 \\
+ SIMA$^{\dag}$ \cite{wang2024enhancing} &60.1	&32.4	&2.11	&0.55	&6.4	&47.4	&26.1	&3.2	&73.4	&76.4	&\underline{65.0} \\
+ RLAIF-V$^{\dag}$ \cite{yu2024rlaif}&62.8	&29.2	&\textbf{2.95}	&\textbf{0.34}	&\textbf{2.9}	&50.2	&\textbf{16.0}	& \textbf{1.0}	&54.2	&73.7	&63.5 \\
+ CSR$^{\dag}$ \cite{zhou2024calibrated}&\underline{65.7}	&32.2	&2.07	&0.60	&\underline{3.8}	&45.0	&\underline{16.9}	&\underline{1.4}	&73.1	&76.0	&64.1 \\
\midrule
\multicolumn{6}{l}{\textit{without annotated data or extra models:}} \\
\midrule
 + SENA (Ours) &\textbf{67.4}	&\underline{35.8}	&\underline{2.33}	&\underline{0.52}	&4.9	&49.4	&20.5	&1.7	&\textbf{79.4}	&\textbf{83.6}	&64.8 \\
 \hline

\end{tabular}
\caption{Comparisons with multiple MLLMs and various self-evolution frameworks. $\dag$ indicates evaluation results based on the models released by the authors, while $\dag\dag$ indicates evaluation results based on the code released by the authors.}
\label{tab:sota}
\end{table*}

\subsection{Evaluation Benchmarks}
We evaluate our model's generative and discriminative capabilities. Generative capability assesses the model's ability to produce detailed and accurate content, using benchmarks such as LLaVA$^{\mathrm{W}}$ \cite{liu2024visual}, MM-Vet \cite{yu2023mm}, MMHal-Bench \cite{sun2023aligning}, and AMBER \cite{wang2023llm}. Discriminative capability measures the model's ability to distinguish between relevant and irrelevant information, leveraging benchmarks like AMBER \cite{wang2023llm} and MMBench \cite{liu2023mmbench}. Detailed descriptions of these benchmarks are available in the Supplementary Materials.
Some benchmarks require scoring using the GPT-4 API. To ensure fair comparisons, we use the GPT-4-1106-preview version for all tests, as there can be significant performance variability across different API versions. Each test will be conducted three times, with the average score taken as the final result to minimize randomness and ensure reliable evaluation.

\begin{figure*}[ht]
\centerline{\includegraphics[width=1.0\linewidth]{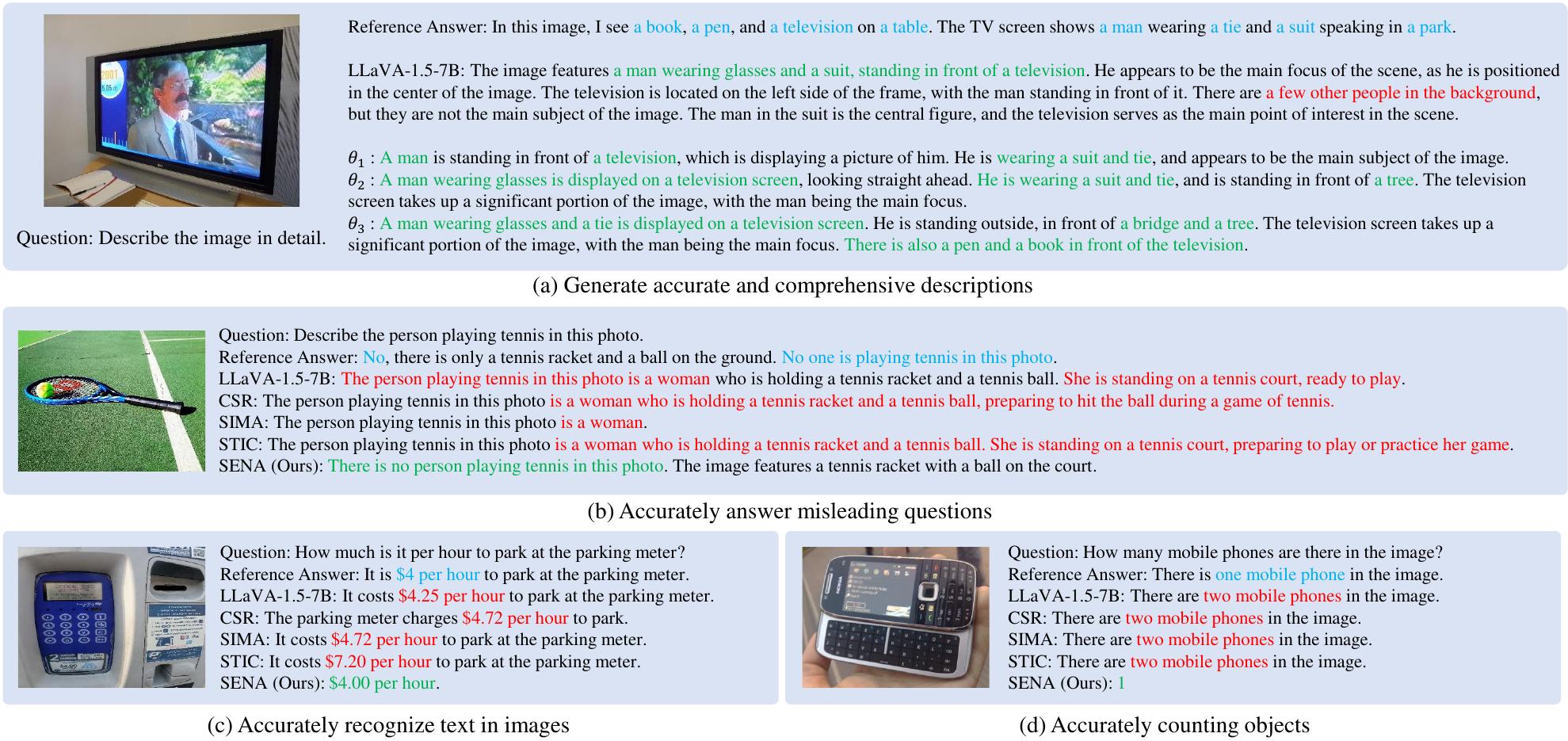}}
  \caption{Comparison of outputs from various models on different visual tasks in MMHal-Bench.
Best viewed in color.
}
  \label{fig:figure_visual}
\end{figure*}

\subsection{Ablation Study}
We conduct a comprehensive ablation study using the LLaVA-1.5-7B ($\theta_0$) model. Initially, $\theta_0$ undergoes one evolution round under the baseline framework, resulting in $\theta_{1}^{Base}$. We then add methods SQ, SE, and CA individually to the baseline, each undergoing one evolution round, creating $\theta_1^{SQ}$, $\theta_1^{SE}$, and $\theta_1^{CA}$, respectively. This allows us to assess each component separately. Lastly, we combine all methods into a complete framework and conduct three evolution rounds starting from $\theta_0$, resulting in models $\theta_1$, $\theta_2$, and $\theta_3$. All results are listed in Table \ref{tab:ablation_study}.

\textbf{(a) Baseline Framework:} The baseline framework allows the model to self-generate questions and answers without labeled data or extra models. It improves some benchmarks, but $\theta_{1}^{Base}$ struggles with hallucination-related benchmarks like MMHal-bench due to noise in the generated data, indicating a need for content refinement. 
\textbf{(b) Self-Questioning:} The SQ mechanism refines the quality of generated questions by identifying and regenerating meaningless ones. The performance of $\theta_1^{SQ}$ shows noticeable improvement over $\theta_1^{Base}$.
\textbf{(c) Self-Enhancement:} SE enhances the quality of the chosen answers utilizing image description. The model $\theta_1^{SE}$ achieves significant progress compared to $\theta_1^{Base}$, while also broadly enhancing the generation and recognition abilities of $\theta_0$. This indicates that SE effectively helps the model learn valuable knowledge from the generated data.
\textbf{(d) Content Alignment:} CA is designed to enhance the model’s focus on image content, greatly improving the model's generative ability and discriminative power. For example, $\theta_1^{CA}$ achieves a high score of 66.7 in LLaVA$^{\mathrm{W}}$ and an excellent F1 score of 77.9 in the AMBER-Discriminative task.
\textbf{(e) Overall Framework Effectiveness:} 
First, the performance of $\theta_1$ clearly exceeded that of the previous models, indicating that the three key designs are complementary. Furthermore, the performance of the model evolved in each iteration generally improves compared to the previous iteration, and all three models significantly outperform the $\theta_0$ model. This strongly validates the effectiveness of our framework. Notably, all these enhancements are achieved using unlabeled images, highlighting the scalability and great practical value of our approaches.

\textbf{Impact of SE on Answers.}
SE uses image descriptions to enhance the quality of all chosen answers. As shown in Table \ref{tab:si_term}, even improving just one type of answer, such as $y_{gen}$ or $y_{des}$, can boost model performance. This clearly demonstrates the effectiveness of SE. \textbf{Impact of CA on Answers.} CA enhances the model's ability to focus on images by maximizing the log-likelihood of descriptive answers, thereby reducing hallucinations. This loss can also be applied to generated answers. As shown in Table \ref{tab:sa_term}, applying \(L_{Align}\) to generated answers yields positive results as well. However, its performance is optimal when applied to descriptive answers. This is because descriptive answers encompass most of the image content, while generated answers may only focus on specific objects within the image.

\subsection{Comparison with SOTA}
We name the model \(\theta_3\) as SENA and compare it with other models. The comparison results are summarized in Table \ref{tab:sota}.

Existing methods heavily rely on annotations. For instance, SeVa \cite{zhu2024self} uses instruction data from TextVQA and OCRVQA to improve the model's OCR capabilities. However, SeVa focuses on constructing hard negative rejected answers without addressing potential hallucination issues, resulting in a notable hallucination problem. Additionally, some methods require the ground truth answers to the questions. STIC \cite{deng2024enhancing} mixes these ground truth answers with generated data for supervised fine-tuning, while SIMA \cite{wang2024enhancing} uses them for answer selection, leading to significant annotation costs.

Some methods also necessitate extra models. RLAIF-V \cite{yu2024rlaif} shows strong anti-hallucination performance by using the 34B LLaVA-NEXT model to filter responses from the 7B LLaVA-1.5 model. However, this can lead to a preference for shorter, less detailed responses, affecting performance on generative benchmarks like MM-VET. CSR \cite{zhou2024calibrated} uses CLIP scores to select answers, with higher-scoring responses being chosen and lower-scoring ones being rejected. This approach improves the model's performance on image captioning tasks. Take the examples shown in Fig. \ref{fig:figure_se}, when a query involves multiple objects in an image, a response like ``giraffe and zebras" is more precise than just ``giraffe," resulting in a higher CLIP score. However, there are instances where CLIP scores can be misleading. For questions targeting a single object, CLIP tends to favor responses that mention multiple objects, which may not align with the actual query. Consequently, CSR might sometimes generate answers that do not match the instructions, negatively affecting its ability to accurately follow instructions on tasks like MM-VET.

In contrast, SENA uses only unlabeled images to generate unique questions, combined with general descriptive questions to provide diverse instructions. It also introduces three technical solutions to improve question generation, answer generation, and preference optimization, thereby enhancing the model's generative and discriminative abilities while maintaining strong resistance to hallucinations.

\subsection{Qualitative analysis}
In this section, we conduct a qualitative analysis of our model's evaluation results on MMHal-Bench \cite{sun2023aligning}, exploring which specific capabilities of the model have been enhanced during the self-evolution process. The results are presented in Fig. \ref{fig:figure_visual}.

\textbf{(a) Generate Accurate and Comprehensive Descriptions.}
The LLaVA-1.5-7B model often provides detailed descriptions but may lead to hallucinations, such as incorrectly mentioning other people in the image. In contrast, our models deliver accurate descriptions. Notably, as self-evolution progresses, our model increasingly focuses on finer details in the images. For instance, $\theta_2$ shows greater attention to the trees in the image compared to $\theta_1$. Furthermore, the $\theta_3$ model not only accurately describes the man but also notes a pen and a book in front of the television, showcasing its improved focus on key elements. This enhancement is attributed to our CA loss, which helps the model better attend to the content of the images.
\textbf{(b) Accurately Answer Misleading Questions.}
Some models struggle with misleading questions as they inaccurately describe a woman playing tennis despite the reference indicating no one is playing. Conversely, the SENA model correctly notes the absence of players and mentions a racket and ball, demonstrating better interpretation of misleading questions.
\textbf{(c) Accurately Recognize Text in Images.}
The SENA model excels in text recognition. For example, when asked about the hourly parking fee, LLaVA-1.5-7B states \$4.25, while SENA accurately identifies it as \$4.00, aligning with the reference answer. \textbf{(d) Accurately Counting Objects.} Accurate object counting is crucial for visual understanding. The LLaVA-1.5-7B miscounts two phones in the image, while the SENA correctly identifies a single phone, highlighting its enhanced ability to locate and count objects in images.

\section{Conclusion and Limitations}
This paper introduces SENA, a multi-model self-evolution framework that differs significantly from traditional methods as it does not require arbitrary annotations. This framework is supported by three mechanisms: image-driven self-questioning, answer self-enhancement, and an image content alignment function. These mechanisms address key challenges in generating reliable questions, constructing discriminative preferences data, and optimizing the model to reduce hallucinations. Experimental results and qualitative analysis indicate that the SENA model significantly outperforms the baseline model across various tasks, excelling in generating accurate descriptions, answering misleading questions, recognizing text in images, and counting objects. 

However, our framework still has limitations. For instance, its performance on certain benchmarks lags behind self-evolution methods that utilize annotated data, and there is a performance plateau after three rounds of evolution. These challenges motivate our ongoing efforts to refine and improve the framework in future work.

\textbf{Acknowledgement.} This work was partially supported by the Major Science and Technology Innovation 2030 ``New Generation Artificial Intelligence" key project (No. 2021ZD0111700), the National Natural Science Foundation of China under Grants 62476099 and 62076101, the Guangdong Basic and Applied Basic Research Foundation under Grants 2024B1515020082 and 2023A1515010007, the Guangdong Provincial Key Laboratory of Human Digital Twin under Grant 2022B1212010004, the TCL Young Scholars Program, and the 2024 Tencent AI Lab Rhino-Bird Focused Research Program.

\bibliography{aaai25}

\clearpage

\section{General Descriptive Prompt Set $P_{des}$}
\begin{figure}[ht]
\centerline{\includegraphics[width=1.0\linewidth]{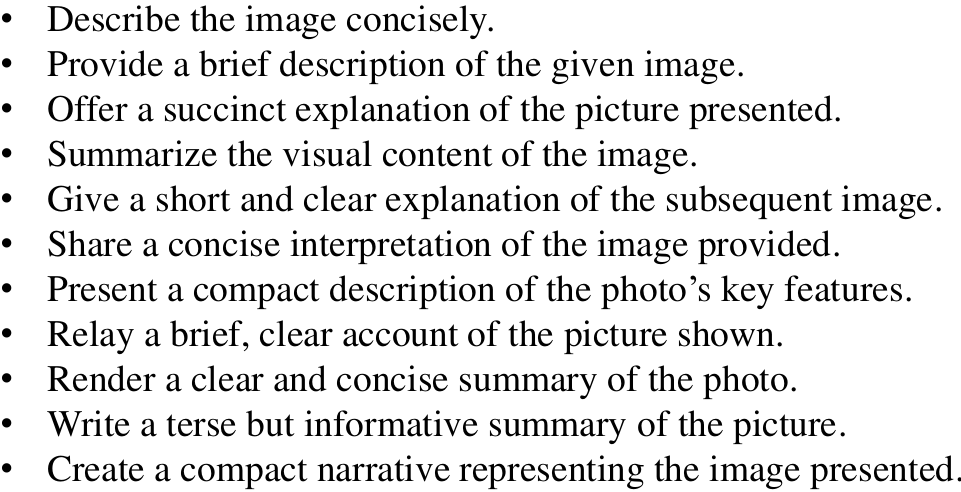}}
  \caption{The General Descriptive Prompt Set $P_{des}$.
}
  \label{fig:figure_pdes}
\end{figure}

SENA enables the model to generate questions 
$q_{gen}$ about images. We find that $q_{gen}$ sometimes focuses only on a specific object in the image, neglecting other important information. To maximize the knowledge the model learns from the image, we add a descriptive question $q_{des}$ for each image. This $q_{des}$ is randomly sampled from the general descriptive prompt set $P_{des}$, which is adapted from \cite{liu2024visual}, and is illustrated in Fig. \ref{fig:figure_pdes}. Obtaining image descriptions also provides a foundation for answer self-enhancement technique and image content alignment loss, making descriptive questions an important part of our framework.

\section{The Issue of Similar Quality in Answers}
\begin{figure}[ht]
\centerline{\includegraphics[width=1.0\linewidth]{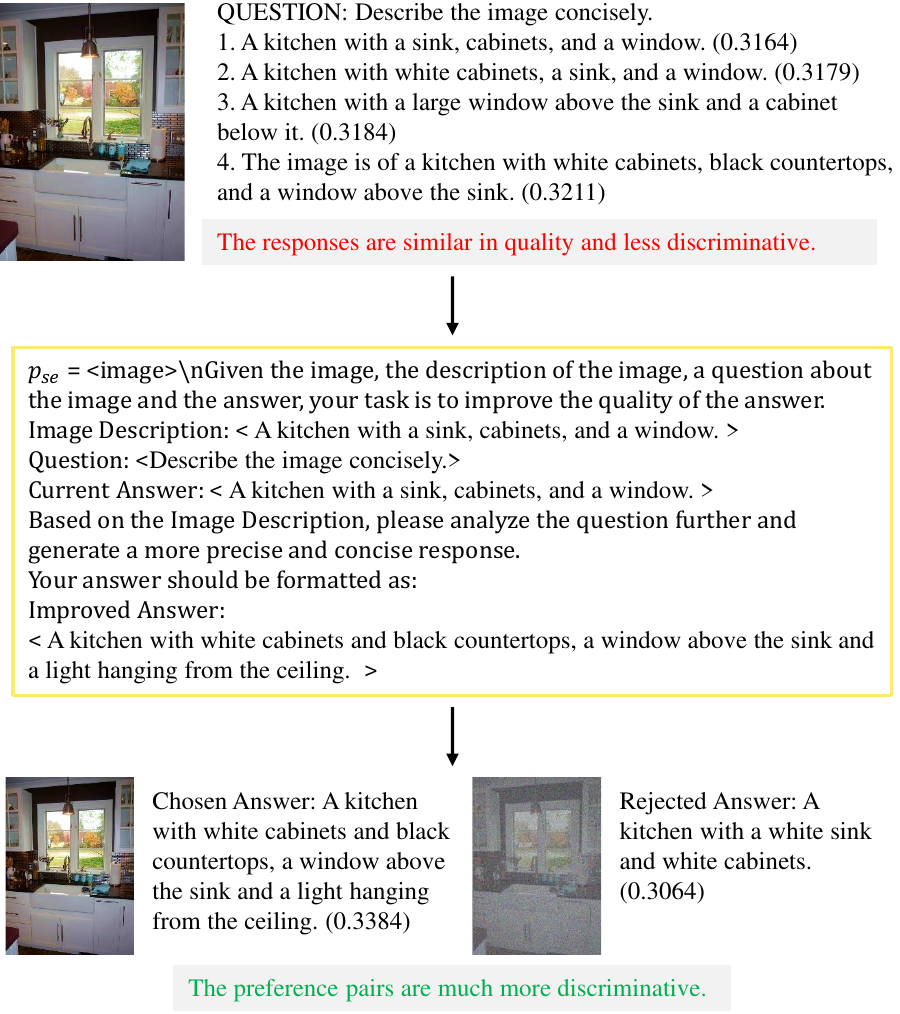}}
  \caption{The General Descriptive Prompt Set $P_{des}$.
}
  \label{fig:figure_similar}
\end{figure}
The quality of randomly generated model answers is often similar, as shown in Fig. \ref{fig:figure_similar}. This results in insufficient discriminative power when selecting the best and worst answers for preference data. Although some methods use augmented images to generate rejected answers \cite{deng2024enhancing,zhu2024self}, these can still be similar to the chosen responses.

Our Answer Self-Enhancement approach improves the quality of the chosen answers. For instance, the enhanced chosen answer achieves a CLIP score of 0.3384, higher than the original score of 0.3164. This improvement arises from a more detailed description that includes elements like ``a light hanging from the ceiling." Such enhancements increase the discriminative power between preferences, facilitating more effective human preference alignment.

\section{Benchmark Details}
\begin{itemize}
\item \textbf{LLaVA$^{\mathrm{W}}$ \cite{liu2024visual}}  

LLaVA$^{\mathrm{W}}$ is a comprehensive benchmark for evaluating MLLMs. It consists of 24 images covering various scenes, along with 60 questions that assess the models' capabilities in dialogue, description, and reasoning. The model's performance is measured as the ratio of the score for its responses to the score for the reference answers, with the response scores evaluated by GPT-4 \cite{achiam2023gpt}.

\begin{table*}[htp]
\small
\centering
\begin{tabular}{l||cccc|cccc|cccc}
\toprule[1pt]
\multirow{2}{*}{GPT Version} & \multicolumn{4}{c|}{LLaVA$^{\mathrm{W}}$} & \multicolumn{4}{c|}{MM-Vet} & \multicolumn{4}{c}{MMHal-Bench} \\ \cline{2-13} 
 & 1 & 2 & 3 & Avg. & 1 & 2 & 3 & Avg. & 1 & 2 & 3 & Avg. \\ \hline \hline
GPT-3.5-turbo-0125 &86.3  &85.3  &82.8  &84.8  &33.2  &32.3  &32.4  &32.6  &3.36/0.32  &3.31/0.31  &3.40/0.30  &3.36/0.31  \\
GPT-4-1106-preview &60.7  &59.7  &58.4  &59.6  &31.3  &32.2  &31.6  &31.7  &1.89/0.59  &1.96/0.58  &1.86/0.66  &1.90/0.61  \\
GPT-4o &57.8  &57.0  &58.2  &57.7  &28.3  &28.1  &28.1  &28.2  &1.91/0.66  &1.95/0.65  &1.92/0.65  &1.93/0.65  \\
\hline
\end{tabular}
\caption{Performance comparison of the LLaVA-1.5-7B model using different GPT API versions across various benchmarks: LLaVA$^{\mathrm{W}}$, MM-Vet, and MMHal-Bench. Each cell shows the scores for three test cases, along with their average scores. }

\label{tab:api}
\end{table*}

\item \textbf{MM-Vet \cite{yu2023mm}} 

MM-Vet comprises 200 images and 218 questions aimed at assessing six core vision-language capabilities: recognition, OCR, knowledge, language generation, spatial reasoning, and mathematics. The evaluation process utilizes GPT-4 \cite{achiam2023gpt}, which is prompted with few-shot evaluation examples to generate scores for the model's responses, ranging from 0 to 1. The overall performance of the model is determined by the sum of all these scores.

\item \textbf{MMHal-Bench \cite{sun2023aligning}}

MMHal-Bench consists of 96 image-question pairs and utilizes GPT-4 \cite{achiam2023gpt} to analyze and score the model's responses. Specifically, GPT-4 compares the model's answers to the standard human-generated reference responses, assigning quality scores and determining the presence of hallucinations. The overall performance of the model is calculated as the average of all scores (Score), along with the rate of hallucinations (Rate$\downarrow$).

\item \textbf{AMBER \cite{wang2023llm}}

AMBER comprises a total of 1,004 images, each annotated with four 4 types of content:
\begin{itemize}
    \item \textit{Existence} refers to all visible objects in the image.
    \item  \textit{Attribute} describes the characteristics of existing objects, including state (such as color and shape), number, and action.
    \item \textit{Relation} indicates whether there is direct contact between two objects in the image.
    \item \textit{Hallucinatory target objects} that are explicitly absent from the image but may be imagined by the MLLMs.
\end{itemize}
All annotations are verified by human reviewers to ensure reliability.

AMBER is designed to handle both generative and discriminative tasks. For generative tasks, AMBER utilizes the common prompt ``Describe this image" to obtain descriptions from the MLLM. For discriminative tasks, AMBER creates specific prompts based on the type of hallucination. These prompts typically start with ``Is," ``Are," or ``Does," and ask whether the elements in the question align with the annotations provided for the image. The model simply needs to respond with "Yes" or "No."

AMBER includes several metrics. For generative task: 

\begin{itemize}
    \item \textit{CHAIR$\downarrow$} quantifies the occurrence of hallucinated objects in the model’s responses, defined as: 
      \[
  CHAIR(R) = 1 - \frac{len(R'_{obj} \cap A_{obj})}{len(R'_{obj})},
  \]
  where $R$ represent one model's response, $R'_{obj}$ is the set of all object nouns present in the $R$ (after applying reasonable filtering), and $A_{obj}$ represents the \textit{Existence} annotations for the image. If the model’s response includes objects that are not part of the image annotations, it indicates a hallucination, resulting in a higher \textit{CHAIR} score.

  \item \textit{Cover} measures the object coverage of responses, quantifying the proportion of objects mentioned in the response \( R'_{obj} \) relative to those identified in \( A_{obj} \). \textit{Cover} is calculated as:
  \[
  Cover(R) = \frac{len(R'_{obj} \cap A_{obj})}{len(A_{obj})}.
  \]

  \item \textit{Hal$\downarrow$} indicates the fraction of responses that contain hallucinations. It is defined as:
  \[
  Hal(R) = 
  \begin{cases} 
  1 & \text{if } CHAIR(R) \neq 0 \\
  0 & \text{otherwise}
  \end{cases}.
  \]

  \item \textit{Cog$\downarrow$} evaluates whether the hallucinations produced by MLLMs are similar to those found in human cognition:
  \[
  Cog(R) = \frac{len(R'_{obj} \cap H_{obj})}{len(R'_{obj})},
  \]
  where $H_{obj}$ is a set of human hallucination target objects.
\end{itemize}

For discriminative task, MLLMs must respond with a ``Yes" or ``No". AMBER employs standard classification metrics—accuracy, precision, recall, and F1 score—to evaluate the model's discriminative ability. In our experiments, we only record accuracy and the F1 score.

\item \textbf{MMBench \cite{liu2023mmbench}}

MMBench includes approximately 3K multiple-choice questions that cover 20 different capability dimensions of MLLMs, such as object localization and logical reasoning. Each capability dimension contains over 75 questions, allowing for a balanced and comprehensive evaluation of the model. To enhance the robustness of the assessment, MMBench introduces a strategy called CircularEval. In this approach, the options are shuffled multiple times when asking questions to the MLLM. The MLLM must provide correct answers across all variations of the same question for its response to be deemed truly correct. Finally, the model's performance is represented by the average score.

\end{itemize}

\section{More Experiments}

\subsection{The Impact of GPT Version}
The benchmarks LLaVA$^{\mathrm{W}}$ \cite{liu2024visual}, MM-Vet \cite{yu2023mm} and MMHal-Bench \cite{sun2023aligning} utilize GPT-4 for scoring. In this section, we will utilize the GPT-4-1106-preview mentioned in the main paper, along with the GPT-3.5-turbo-0125, and the more advanced GPT-4o API version, to test the LLaVA-1.5-7B model. This will allow us to evaluate how different API versions affect the model's performance.

The results in the Table \ref{tab:api} indicate that the performance of the LLaVA-1.5-7B model is significantly influenced by the version of the GPT API used. Newer versions of the GPT model tend to be more stringent in their evaluations, resulting in lower scores for the model’s responses. Additionally, there can be fluctuations in the results when using the same GPT multiple times. Therefore, in our main text, we first specify the GPT version used and conduct three test runs, taking the average of these results as the final reported performance of the model.

\subsection{The Impact of Image Size $M$}
SENA uses $M$ images in each iteration and the impact of varying $M$ on performance is summarized in Table \ref{tab:M}. The table shows that as the number of images increases, the overall performance of the model improves. This improvement occurs because the model gains more knowledge with more images, leading to better performance. However, beyond a certain threshold, such as 6K images, the performance no longer shows significant improvement. This is because even though SQ and SE are employed to enhance the quality of the generated data, there is still a small probability that the model may produce hallucinated samples. As the number of images grows, the number of hallucinated samples also increases, eventually causing the performance to plateau. Nonetheless, it is evident that the performance of the model using SENA is significantly better compared to the baseline model overall.

\subsection{The Impact of Diffusion Noise Steps $T$} 
In SENA, rejected responses are generated using $x'$, which is an image $x$ processed with diffusion noise. The hyperparameter $T$ controls the level of noise added. By adjusting $T$, we produce rejected responses of varying quality. The impact of $T$ are shown in Table \ref{tab:T}. 

As $T$ increases, the model's performance improves. It is because larger $T$ makes poorer responses, creating strongly discriminative preference dataset when combined with enhanced chosen answers. However, once $T$ exceeds a certain threshold, the model can barely extract any useful information from the images. As a result, the rejected answers become less challenging, and performance begins to decline. In summary, the optimal value for $T$ is around 600.

\subsection{Diversifying Questions is Beneficial}
SENA equips each image with two types of questions: model-generated questions ($q_{gen}$) and descriptive questions ($q_{des}$). It is anticipated that the model's inherent biases may lead to a fixed focus on certain themes when generating $q_{gen}$. On the other hand, relying solely on $q_{des}$ results in a significant lack of diversity. We conduct an experiment to validate this hypothesis, and the results are presented in Table \ref{tab:question}. The findings indicate that while using either type of question alone can improve the model's performance beyond the baseline, the best results are achieved only when both types are combined.

In fact, the diversity of questions has been shown to aid model learning across various fields. Self-evolution, which heavily relies on model-generated content for training, must not only address issues of hallucination but also prioritize the study of diversity. We plan to explore this aspect in our future work.

\begin{table}[t]
    \centering
    \caption{Model Performance on Common VQA Benchmarks}
    \begin{tabular}{lcccc}
        \toprule
        Model & GQA & SQA & TextVQA & VQAv2 \\
        \midrule
        LLaVA-1.5-7B & 62.0 & 70.4 & 46.1 & 76.6 \\
        SeVa & 61.6 & 69.1 & 42.9 & 76.4 \\
        STIC & 58.7 & 64.7 & 47.3 & 76.0 \\
        SIMA & 62.8 & 69.7 & 46.9 & 76.6 \\
        RLAIF-V & 59.2 & 70.9 & 44.6 & 73.1 \\
        CSR & 62.9 & 69.8 & 46.4 & \textbf{76.8} \\
        SENA (Ours)& \textbf{63.3} & \textbf{71.9} & \textbf{47.8} & 76.4 \\
        \bottomrule
    \end{tabular}
    \label{tab:vqa}
\end{table}

\begin{table*}[t]
\centering
\setlength{\tabcolsep}{1.0mm} 
\small
\begin{tabular}{c||ccc|c||c|c|cc|cccc|cc|c}
\toprule[1pt]
\multirow{3}{*}{$M$} & \multicolumn{3}{c|}{Component} & \multirow{3}{*}{Iteration} & \multicolumn{8}{c|}{Generative Task} & \multicolumn{3}{c}{Discriminative Task} \\ \cline{2-4} \cline{6-16} 
 & \multirow{2}{*}{SQ} & \multirow{2}{*}{SE} & \multirow{2}{*}{CA} &  & \multicolumn{1}{c|}{\multirow{2}{*}{LLaVA$^{\mathrm{W}}$}} & \multicolumn{1}{c|}{\multirow{2}{*}{MM-VET}} & \multicolumn{2}{c|}{MMHal} & \multicolumn{4}{c|}{AMBER-Gen.} & \multicolumn{2}{c|}{AMBER-Dis.} & \multirow{2}{*}{MMBench} \\ 
 &  &  &  &  & \multicolumn{1}{c|}{} & \multicolumn{1}{c|}{} & Score & \multicolumn{1}{c|}{Rate$\downarrow$} & CHAIR$\downarrow$ & Cover & Hal$\downarrow$ & Cog$\downarrow$ & Accuracy & \multicolumn{1}{c|}{F1} &  \\ \hline 
- &&&&&59.6 & 31.7 & 1.90 & 0.61 & 7.6 & 51.8 & 35.1 & 4.3 & 71.7  & 74.3 & 64.6 \\ \hline
2K &$\checkmark$&$\checkmark$&$\checkmark$&1&63.6	&32.0	&2.10	&0.56	&6.0	&50.3	&27.0	&2.8	&75.0	&78.0	&64.7 \\
4K &$\checkmark$&$\checkmark$&$\checkmark$&1&64.8	&32.8	&2.16	&0.54	&\textbf{5.2}	&50.6	&\textbf{24.1}	&2.4 	&\textbf{75.6}	&78.7	&\textbf{65.2}\\ 
6K &$\checkmark$&$\checkmark$&$\checkmark$&1&66.9	&\textbf{34.2}	&\textbf{2.28}	&\textbf{0.54}	&5.6	&\textbf{51.6}	&25.2	&\textbf{1.9}	&75.1	&\textbf{79.8}	&65.0 \\
8K &$\checkmark$&$\checkmark$&$\checkmark$&1&\textbf{67.3}	&33.1	&2.11	&0.57	&5.6	&50.3	&25.5	&2.5	&73.9	&77.0	&64.6 \\ 
\hline
\end{tabular}
\caption{The Impact of Images Size $M$.}
\label{tab:M}
\end{table*}

\begin{table*}[t]
\centering
\setlength{\tabcolsep}{1.0mm} 
\small
\begin{tabular}{c||ccc|c||c|c|cc|cccc|cc|c}
\toprule[1pt]
\multirow{3}{*}{$T$} & \multicolumn{3}{c|}{Component} & \multirow{3}{*}{Iteration} & \multicolumn{8}{c|}{Generative Task} & \multicolumn{3}{c}{Discriminative Task} \\ \cline{2-4} \cline{6-16} 
 & \multirow{2}{*}{SQ} & \multirow{2}{*}{SE} & \multirow{2}{*}{CA} &  & \multicolumn{1}{c|}{\multirow{2}{*}{LLaVA$^{\mathrm{W}}$}} & \multicolumn{1}{c|}{\multirow{2}{*}{MM-VET}} & \multicolumn{2}{c|}{MMHal} & \multicolumn{4}{c|}{AMBER-Gen.} & \multicolumn{2}{c|}{AMBER-Dis.} & \multirow{2}{*}{MMBench} \\ 
 &  &  &  &  & \multicolumn{1}{c|}{} & \multicolumn{1}{c|}{} & Score & \multicolumn{1}{c|}{Rate$\downarrow$} & CHAIR$\downarrow$ & Cover & Hal$\downarrow$ & Cog$\downarrow$ & Accuracy & \multicolumn{1}{c|}{F1} &  \\ \hline 
- &&&&&59.6 & 31.7 & 1.90 & 0.61 & 7.6 & 51.8 & 35.1 & 4.3 & 71.7  & 74.3 & 64.6 \\ \hline
200 &$\checkmark$&$\checkmark$&$\checkmark$&1&64.2	&32.4	&2.22	&0.55	&5.9	&50.5	&27.4	&2.6	&74.8	&79.4	&64.3 \\ 
400 &$\checkmark$&$\checkmark$&$\checkmark$&1&66.8	&33.7	&2.25	&\textbf{0.54}	&\textbf{5.6}	&\textbf{51.8}	&25.3	&2.9	&\textbf{75.3}	&79.7	&64.8 \\
600 &$\checkmark$&$\checkmark$&$\checkmark$&1&\textbf{66.9}	&\textbf{34.2}	&\textbf{2.28}	&\textbf{0.54}	&\textbf{5.6}	&51.6	&\textbf{25.2}	&\textbf{1.9}	&75.1	&\textbf{79.8}	&65.0 \\
800 &$\checkmark$&$\checkmark$&$\checkmark$&1&66.6	&33.6	&2.24	&\textbf{0.54}	&5.7	&51.3	&28.4	&3.0	&73.6	&76.7	&65.2 \\
999 &$\checkmark$&$\checkmark$&$\checkmark$&1&64.5	&32.1	&2.16	&0.55	&5.7	&51.5	&27.6	&2.7	&68.8	&70.5	&\textbf{65.5} \\
\hline
\end{tabular}
\caption{The Impact of Diffusion Noise Steps $T$.}
\label{tab:T}
\end{table*}

\begin{table*}[t]
\centering
\setlength{\tabcolsep}{1.0mm} 
\small
\begin{tabular}{c||ccc|c||c|c|cc|cccc|cc|c}
\toprule[1pt]
\multirow{3}{*}{Question} & \multicolumn{3}{c|}{Component} & \multirow{3}{*}{Iteration} & \multicolumn{8}{c|}{Generative Task} & \multicolumn{3}{c}{Discriminative Task} \\ \cline{2-4} \cline{6-16} 
 & \multirow{2}{*}{SQ} & \multirow{2}{*}{SE} & \multirow{2}{*}{CA} &  & \multicolumn{1}{c|}{\multirow{2}{*}{LLaVA$^{\mathrm{W}}$}} & \multicolumn{1}{c|}{\multirow{2}{*}{MM-VET}} & \multicolumn{2}{c|}{MMHal} & \multicolumn{4}{c|}{AMBER-Gen.} & \multicolumn{2}{c|}{AMBER-Dis.} & \multirow{2}{*}{MMBench} \\ 
 &  &  &  &  & \multicolumn{1}{c|}{} & \multicolumn{1}{c|}{} & Score & \multicolumn{1}{c|}{Rate$\downarrow$} & CHAIR$\downarrow$ & Cover & Hal$\downarrow$ & Cog$\downarrow$ & Accuracy & \multicolumn{1}{c|}{F1} &  \\ \hline 
- &&&&&59.6 & 31.7 & 1.90 & 0.61 & 7.6 & 51.8 & 35.1 & 4.3 & 71.7  & 74.3 & 64.6 \\ \hline
$q_{des}^{sq}$ &$\checkmark$&$\checkmark$& &1&62.7	&33.0	&2.08	&0.57	&6.5	&50.6	&30.6	&3.4	&74.1	&77.2	&64.9 \\
$q_{gen}^{sq}$ &$\checkmark$&$\checkmark$& &1&64.5	&33.1	&2.14	&0.55	&6.5	&\textbf{51.2}	&31.9	&3.4 	&71.9	&74.2	&\textbf{65.5}\\ 
$q_{des}^{sq}$ + $q_{gen}^{sq}$ &$\checkmark$&$\checkmark$& &1&\textbf{65.3}	&\textbf{33.4}	&\textbf{2.24}	&\textbf{0.53}	&\textbf{5.9}	&51.0	&\textbf{29.6}	&\textbf{3.2}	&\textbf{74.3}	&\textbf{77.3}	&65.2 \\
\hline
\end{tabular}
\caption{Performance comparison of the model using different types of questions.}
\label{tab:question}
\end{table*}

\begin{table*}[p]
\centering
\setlength{\tabcolsep}{1.0mm} 
\small
\begin{tabular}{c||ccc|c||c|c|cc|cccc|cc|c}
\toprule[1pt]
\multirow{3}{*}{Method} & \multicolumn{3}{c|}{Component} & \multirow{3}{*}{Iteration} & \multicolumn{8}{c|}{Generative Task} & \multicolumn{3}{c}{Discriminative Task} \\ \cline{2-4} \cline{6-16} 
 & \multirow{2}{*}{SQ} & \multirow{2}{*}{SE} & \multirow{2}{*}{CA} &  & \multicolumn{1}{c|}{\multirow{2}{*}{LLaVA$^{\mathrm{W}}$}} & \multicolumn{1}{c|}{\multirow{2}{*}{MM-VET}} & \multicolumn{2}{c|}{MMHal} & \multicolumn{4}{c|}{AMBER-Gen.} & \multicolumn{2}{c|}{AMBER-Dis.} & \multirow{2}{*}{MMBench} \\ 
 &  &  &  &  & \multicolumn{1}{c|}{} & \multicolumn{1}{c|}{} & Score & \multicolumn{1}{c|}{Rate$\downarrow$} & CHAIR$\downarrow$ & Cover & Hal$\downarrow$ & Cog$\downarrow$ & Accuracy & \multicolumn{1}{c|}{F1} &  \\ \hline 
$\theta_0$ &&&&&66.8 & 36.7 & 2.26 & 0.59 & 6.5 & 52.0 & 30.6 & 3.3	 & 71.2  & 73 & 68.4 \\ \hline
$\theta_1$ &$\checkmark$&$\checkmark$&$\checkmark$&1&\underline{70.5}	&39.8	&\underline{2.30}	&\underline{0.54}	&6.7	&51.9	&30.2	&2.9	&77.1	&80.2	&\underline{68.6} \\

$\theta_2$ &$\checkmark$&$\checkmark$&$\checkmark$&2&69.9	&\underline{40.8}	&2.26	&\underline{0.54}	&\underline{6.3}	&\underline{52.1}	&\underline{30.1}	&\underline{2.8}	&\textbf{78.7}		&\textbf{82.2}	&68.5 \\

$\theta_3$ &$\checkmark$&$\checkmark$&$\checkmark$&3&\textbf{71.8}	&\textbf{41.5}	&\textbf{2.42}	&\textbf{0.51}	&\textbf{5.9}	&\textbf{52.2}	&\textbf{27.7}	&\textbf{2.7}	&\underline{78.2}	&\underline{81.7}	&\textbf{68.8} \\
\hline
\end{tabular}
\caption{The self-evolution results of the LLaVA-1.5-13B model.}
\label{tab:13b}
\end{table*}

\begin{table*}[p]
\centering
\setlength{\tabcolsep}{1.0mm} 
\small
\begin{tabular}{c||ccc|c||c|c|cc|cccc|cc|c}
\toprule[1pt]
\multirow{3}{*}{Method} & \multicolumn{3}{c|}{Component} & \multirow{3}{*}{Iteration} & \multicolumn{8}{c|}{Generative Task} & \multicolumn{3}{c}{Discriminative Task} \\ \cline{2-4} \cline{6-16} 
 & \multirow{2}{*}{SQ} & \multirow{2}{*}{SE} & \multirow{2}{*}{CA} &  & \multicolumn{1}{c|}{\multirow{2}{*}{LLaVA$^{\mathrm{W}}$}} & \multicolumn{1}{c|}{\multirow{2}{*}{MM-VET}} & \multicolumn{2}{c|}{MMHal} & \multicolumn{4}{c|}{AMBER-Gen.} & \multicolumn{2}{c|}{AMBER-Dis.} & \multirow{2}{*}{MMBench} \\ 
 &  &  &  &  & \multicolumn{1}{c|}{} & \multicolumn{1}{c|}{} & Score & \multicolumn{1}{c|}{Rate$\downarrow$} & CHAIR$\downarrow$ & Cover & Hal$\downarrow$ & Cog$\downarrow$ & Accuracy & \multicolumn{1}{c|}{F1} &  \\ \hline 
$\theta_0$ &&&&&86.6 & 59.7 & 3.26 & 0.34 & 6.6 & \textbf{72.2} & 54.0 & 5.2	 & \underline{82.7}  & 85.6 & 79.3 \\ \hline
$\theta_1$ &$\checkmark$&$\checkmark$&$\checkmark$&1&\underline{88.6}	&\underline{63.3}	&\underline{3.31}	&\underline{0.30}	&\textbf{5.6}	&\underline{66.9}	&\underline{34.9}	&\underline{2.6}	&80.5	&\underline{87.2}	&\underline{79.9} \\

$\theta_2$ &$\checkmark$&$\checkmark$&$\checkmark$&2&\textbf{89.2}	&\textbf{63.6}	&\textbf{3.53}	&\textbf{0.28}	&\underline{6.1}	&61.0	&\textbf{32.7}	&\textbf{2.2}	&\textbf{84.3}		&\textbf{89.1}	&\textbf{80.2} \\

\hline
\end{tabular}
\caption{The self-evolution results of the Qwen2-VL-7B model.}
\label{tab:qwen2vl}
\end{table*}

\subsection{Applying SENA on More Base Models}
We apply our self-evolution framework to the LLaVA-1.5-13B model \cite{liu2024llavanext} and more advanced Qwen2-VL-7B model \cite{Qwen2VL}. The experimental results are summarized in Table \ref{tab:13b} and \ref{tab:qwen2vl}. It shows that SENA enhances model's performance in both generative and discriminative tasks. This aligns with the conclusions drawn from the LLaVA-1.5-7B model discussed in the main paper, indicating that our framework is adaptable to various models.

\subsection{More Evaluations on Common VQA Benchmarks}
The VQA benchmarks assess a model's discriminative ability and we use AMBER as a substitute in the paper. For a thorough comparison, we evaluate multiple self-evolution frameworks on common VQA benchmarks using the lmms-eval project \cite{zhang2024lmms}. The results in Table \ref{tab:vqa} confirm SENA's great discriminative ability, which corresponds to the conclusions in Table 4.



\end{document}